\documentclass[10pt]{article}
\usepackage[preprint]{tmlr}

\usepackage{amsmath,amsfonts,bm}

\def\eqref#1{equation~\ref{#1}}
\def\1{\bm{1}}

\DeclareMathAlphabet{\mathsfit}{\encodingdefault}{\sfdefault}{m}{sl}
\SetMathAlphabet{\mathsfit}{bold}{\encodingdefault}{\sfdefault}{bx}{n}

\newcommand{\E}{\mathbb{E}}

\DeclareMathOperator*{\argmax}{arg\,max}

\usepackage{hyperref}
\usepackage{url}
\usepackage{amsmath}
\usepackage{amssymb}
\usepackage{booktabs}
\usepackage{subfig}
\usepackage{float}
\usepackage{algorithm}
\usepackage{algpseudocode}
\usepackage{xcolor}
\usepackage{microtype}
\usepackage{graphicx}

\providecommand{\argmax}{\operatorname*{arg\,max}}\renewcommand{\argmax}{\operatorname*{arg\,max}}

\newcommand{\aspace}{\mathcal{A}}
\newcommand{\sspace}{\mathcal{S}}
\newcommand{\rset}{\mathbb{R}}
\providecommand{\E}{\mathbb{E}}\renewcommand{\E}{\mathbb{E}}

\newcommand{\gqn}{\textsc{GQN}}
\newcommand{\decqn}{\textsc{DecQN}}

\newcommand{\dmc}{DMControl}

\def\month{MM}%
\def\year{YYYY}%
\def\openreview{\url{https://openreview.net/forum?id=XXXX}}%

\begin{document}

\title{Selective Growing Q-Networks (SGQN): An Analysis of Adaptive Action Grids in Decoupled Q-Learning}

\author{\name Parham Nooranbakht \email parham.nooranbakht@email.uni-freiburg.de \\
       \addr Faculty of Engineering, Department of Computer Science \\
       Albert-Ludwigs-Universit\"{a}t Freiburg, Germany
       \AND
       \name Seyed Mahdi Basiri Azad \email basiri@cs.uni-freiburg.de \\
       \addr Faculty of Engineering, Department of Computer Science \\
       Albert-Ludwigs-Universit\"{a}t Freiburg, Germany
       \AND
       \name Joschka B\"{o}decker \email jboedeck@informatik.uni-freiburg.de \\
       \addr Faculty of Engineering, Department of Computer Science \\
       Albert-Ludwigs-Universit\"{a}t Freiburg, Germany}

\maketitle

\begin{abstract}
One way to tackle reinforcement learning for continuous control is to discretise the action space and apply value-based Q-learning, where the resolution of that discretisation shapes the trade-off between exploration and exploitation. Coarse discretisations facilitate early exploration but limit final precision, while fine discretisations enable precise control but slow convergence. Growing Q-Networks (GQN) address this by growing the resolution uniformly from coarse to fine. We introduce \textbf{Selective Growing Q-Networks (SGQN)}, which keeps the decoupled Q-learning framework and replaces uniform growth with an adaptive action grid: bins are added next to bins whose estimated Q-value is high, and bins whose Q-value is low are pruned, with $\varepsilon$-greedy action selection. On eight tasks from the DeepMind Control Suite (DMControl) with 2 to 12 action dimensions, SGQN reaches the highest mean final return among the compared methods on four of them, is within 1.1\% of the best method on three more, and is 3.4\% below Bang-Bang MPO on the remaining one (Walker Run), while using on average 7.5 of the 9 available bins per dimension at the end of training. We then test whether the adaptive grid is responsible for these results, and find that it is not. In our runs, growth activates every candidate bin and behaves like uniform growth, and removing pruning changes the return by at most 1.1\%. Within SGQN, two other design choices matter instead: starting from a coarse grid, since starting with all 9 bins produces individual seeds that collapse, and $\varepsilon$-greedy selection, since an annealed softmax lowers the return by 23\% to 100\%. SGQN is insensitive to its growth and pruning hyperparameters. SGQN also differs from the GQN and DecQN baselines in other respects, notably one gradient update per environment step instead of one per eight, and we did not control for these, so we do not attribute its advantage to a single component. We therefore present SGQN as a competitive value-based method and, above all, as an analysis of which ingredients of a growing action grid matter and which do not.
\end{abstract}


\section{Introduction}
\label{sec:introduction}

Reinforcement learning for continuous control is commonly approached through actor-critic methods, which directly parameterise a policy over a continuous action space \citep{lillicrap2016continuous, haarnoja2018soft, fujimoto2018addressing, schulman2017proximal}. While effective, these methods introduce substantial complexity: heuristics for training stabilisation, increased compute requirements, and wide hyperparameter search spaces \citep{seyde2022solving}. An alternative is action space discretisation, where Q-learning methods, which avoid an explicit policy parameterisation, are equipped with a suitable discretisation of the action space.

Recent work has shown that this is possible. \citet{seyde2021bangbang} showed that Bernoulli (bang-bang) policies are competitive on many locomotion tasks, drawing on Pontryagin's maximum principle to explain why boundary actions are frequently near-optimal. Building on this, \citet{seyde2022solving} demonstrated that Decoupled Q-Networks (\decqn), which decompose the Q-function across action dimensions and apply bang-bang two-bin discretisation, can match state-of-the-art actor-critic performance on standard benchmarks. \citet{seyde2024growing} extended this line of work with Growing Q-Networks (\gqn), which progressively refine a uniform discretisation from coarse to fine resolution over the course of training.

Despite this progress, \gqn\ activates new bins \emph{uniformly} across the action range, regardless of which regions have been visited or which the learned Q-function considers promising. This motivates the question of whether the resolution of the grid should instead follow the learned value function.

We propose \textbf{Selective Growing Q-Networks (SGQN)}, which is designed to make the active action grid follow the learned values through two mechanisms and uses $\varepsilon$-greedy action selection:

\begin{enumerate}
\item \textbf{Value-guided selective growth.} At each growth step, SGQN identifies bins whose running mean Q-value is close to the best bin of the same dimension and activates their neighbours.

\item \textbf{Lazy pruning.} At each pruning step, bins whose running mean Q-value lies below a quantile of the mature bins of that dimension are deactivated, subject to eligibility conditions that protect new bins, the boundary and the centre of the range, and a minimum number of active bins.
\end{enumerate}

SGQN retains full compatibility with the decoupled Q-learning framework of \decqn\ and \gqn\ \citep{seyde2022solving, seyde2024growing}, benefiting from its linear-time complexity in the number of action dimensions and bins, and uses standard off-policy training with prioritised experience replay, double Q-learning and twin critics.

We evaluate SGQN on eight tasks from the DeepMind Control Suite \citep{tassa2018deepmind} with 2 to 12 action dimensions, against five baselines. SGQN reaches the highest mean final return on four of the eight tasks, is within 1.1\% of the best method on three more, and trails Bang-Bang MPO by 3.4\% on Walker Run. Because the central design claim is that resolution is placed where it matters, we also examine directly where SGQN places resolution (Section~\ref{sec:allocation}) and test each component in ablations (Section~\ref{sec:ablation}). We report what we find, including results that weaken the original motivation: the adaptive grid is not what drives the performance.

\paragraph{Contributions.} This paper makes the following contributions:
\begin{enumerate}
\item We propose SGQN, an adaptive-grid variant of decoupled Q-learning with selective growth, lazy pruning and $\varepsilon$-greedy selection, and evaluate it on eight \dmc\ tasks against five baselines. It has the highest mean return on four tasks and is close to the best method on three more.
\item We analyse the learned action grids: which bins are active over training, how the Q-values of active bins compare to those of inactive bins, and which bins are pruned. In our runs the growth rule activates all candidate bins and the retained and removed bins have nearly identical Q-values, while pruning is weakly but consistently biased towards low-Q bins.
\item We study the design through component ablations and a hyperparameter sensitivity study (Section~\ref{sec:ablation}). Within SGQN, the results attribute the performance to the coarse start and to $\varepsilon$-greedy selection, not to value-guided growth or pruning, and show that the growth and pruning hyperparameters can be varied widely without changing the return.
\end{enumerate}

\section{Related Work}
\label{sec:related_work}

\subsection{Value-Based Methods for Continuous Control}

Q-learning \citep{watkins1992q} and its deep extension DQN \citep{mnih2015human} are natively suited to discrete action spaces. Extending them to continuous actions requires either approximating the argmax over a continuous domain or discretising the space. \citet{gu2016continuous} proposed Normalised Advantage Functions (NAF), which represent the Q-function in a form that admits closed-form continuous maximisation. \citet{tavakoli2018action} introduced action branching, treating each action dimension as an independent head of a shared network, enabling Q-learning in factored multi-dimensional spaces with linear rather than exponential complexity.

\citet{tang2020discretising} showed that naive uniform discretisation of continuous action spaces can match policy gradient methods on several tasks when combined with standard Q-learning improvements, challenging the assumption that continuous action representations are necessary. Building on this, \citet{seyde2022solving} proposed \decqn, combining bang-bang (two-bin) discretisation with full value decomposition across dimensions. By framing each action dimension as a cooperative agent in a multi-agent RL formulation \citep{sunehag2018value, rashid2018qmix}, \decqn\ achieves linear-time complexity in the number of action dimensions $d$ and bins $n$ per dimension, and competitive performance with SAC and TD3 on standard benchmarks.

\subsection{Adaptive Action Space Resolution}

The trade-off between exploration efficiency and precision is well documented in continuous control. Coarse discretisations reduce coordination complexity and facilitate exploration \citep{seyde2022solving}, while fine discretisations enable precise steady-state behaviour. Several approaches have aimed to get the best of both worlds.

\citet{seyde2024growing} proposed \gqn, which grows the action space resolution from coarse to fine based on a stagnation criterion: growth is triggered when the moving-average return falls below a threshold $G_{\text{threshold},t} = (1 - 0.05\,\mathrm{sign}(\mu_\mathrm{MA}))\,\mu_\mathrm{MA} + 0.90\,\sigma_\mathrm{MA}$, where $\mu_\mathrm{MA}$ and $\sigma_\mathrm{MA}$ are the moving-average mean and standard deviation of recent returns. Growth follows a predefined sequence (e.g., $2 \to 3 \to 5 \to 9$ bins), activating new bins uniformly and warm-starting their Q-values through the shared network trunk.

\citet{seo2024continuous} proposed Coarse-to-Fine Q-Networks (CQN), which perform hierarchical refinement \emph{within each action selection}: at each of $L$ levels, the agent selects a bin from a $B$-bin discretisation of the current interval, then zooms into that bin for the next level. CQN achieves resolution $B^L$ with only $L \times B$ Q-value evaluations, but errors compound across levels and early mistakes cannot be corrected.

Our work differs from both \gqn\ and CQN. Relative to \gqn, SGQN replaces uniform growth with a value-guided expansion rule and adds lazy pruning, with the aim of concentrating resources on promising regions rather than distributing them uniformly. Section~\ref{sec:allocation} examines to what extent this happens in practice. Relative to CQN, SGQN maintains a persistent global discretisation that is refined over training, rather than performing per-step hierarchical refinement.

\subsection{Adaptive Discretisation in Other Contexts}

Adaptive discretisation has a long history in function approximation and Monte Carlo methods. In state space, adaptive tile coding \citep{sutton2018reinforcement} allocates resolution where the value function varies most rapidly. In tree search, progressive widening \citep{coulom2007computing} grows the branching factor adaptively. In control, curriculum learning approaches \citep{czarnecki2018mix} start with easier task formulations and progressively increase difficulty. SGQN applies this philosophy specifically to the \emph{action space discretisation} in deep Q-learning.

\subsection{Bang-Bang and Boundary-Seeking Policies}

\citet{seyde2021bangbang} investigated the empirical observation that trained continuous control agents frequently saturate at action boundaries, drawing theoretical support from optimal control theory: Pontryagin's maximum principle implies that minimum-time controllers are bang-bang. This motivates the use of binary (Bernoulli) action discretisation, which they show achieves competitive performance on locomotion tasks despite its extreme coarseness. SGQN begins from this insight and is designed to let resolution grow selectively where the Q-function indicates that finer discrimination is valuable.

\section{Background}
\label{sec:background}

\subsection{Markov Decision Processes and Q-Learning}

We consider an agent interacting with a Markov Decision Process (MDP) defined by the tuple $(\sspace, \aspace, \mathcal{P}, \mathcal{R}, \gamma)$, where $\sspace$ is the state space, $\aspace \in \rset^d$ is the $d$-dimensional continuous action space, $\mathcal{P}(s' \mid s, a)$ is the transition kernel, $\mathcal{R}(s, a) \in \rset$ is the reward function, and $\gamma \in [0, 1)$ is the discount factor. The agent seeks a policy $\pi : \sspace \to \aspace$ maximising the expected discounted return $\E_\pi[\sum_{t=0}^\infty \gamma^t r_t]$.

The optimal action-value function satisfies the Bellman optimality equation:
\begin{equation}
Q^*(s, a) = \E\bigl[r + \gamma \max_{a'} Q^*(s', a') \mid s, a\bigr].
\label{eq:bellman}
\end{equation}
Deep Q-Networks \citep{mnih2015human} approximate $Q^*$ with a neural network $Q(s, a; \theta)$, minimising the TD loss:
\begin{equation}
\mathcal{L}(\theta) = \E_{(s,a,r,s') \sim \mathcal{D}}\bigl[(y_t - Q(s, a; \theta))^2\bigr],
\end{equation}
where the target uses a periodically updated target network $\theta^-$. Double DQN \citep{vanHasselt2016deep} uses the online network to select actions but the target network to evaluate them, reducing overestimation bias:
\begin{equation}
a^* = \argmax_{a'} Q(s_{t+1}, a';\, \theta), \qquad
y_t = r_t + \gamma\, Q(s_{t+1}, a^*;\, \theta^{-}).
\label{eq:double_q}
\end{equation}

\subsection{Decoupled Q-Networks}

For a $d$-dimensional action space discretised into $n$ bins per dimension, the joint action space has size $n^d$, which is intractable for large $d$. \citet{seyde2022solving} propose decomposing the Q-function across dimensions:
\begin{equation}
Q_{\mathrm{total}}(s, \mathbf{a}) = \frac{1}{d} \sum_{i=1}^{d} Q_i(s, a^i),
\label{eq:decqn}
\end{equation}
where $Q_i(s, a^i)$ evaluates only dimension $i$ of the action. Greedy action selection then decouples:
\begin{equation}
a^{*i} = \argmax_{a^i \in \mathcal{A}^i} Q_i(s, a^i), \quad i = 1, \ldots, d,
\end{equation}
reducing complexity from $O(n^d)$ to $O(d \times n)$. The decomposition corresponds to the Individual Global Max (IGM) property from cooperative multi-agent RL \citep{sunehag2018value}, which holds exactly when action dimensions are independent and approximately otherwise.

\subsection{Growing Q-Networks}

\gqn\ \citep{seyde2024growing} maintains a binary active mask $m \in \{0,1\}^{d \times N}$ over a full grid of $N$ bins per dimension, initially activating only $n_0 \ll N$ uniformly-spaced bins. Growth is triggered by a stagnation criterion and advances the active set through a predefined sequence $n_0 \to n_1 \to \cdots \to N$. All newly activated bins are warm-started via the shared network trunk, exploiting the implicit structure learned at coarser resolution. At each step, Q-values are computed and actions selected only over the currently active bins.

\section{Selective Growing Q-Networks}
\label{sec:method}

SGQN builds on the \decqn\ and \gqn\ frameworks. It keeps their decoupled value function and replaces the predefined growth schedule with an adaptive active set $m^i$ per dimension that is changed by two rules, growth and pruning, both driven by running estimates of the Q-value of each bin. Action selection uses $\varepsilon$-greedy exploration over the active bins.

\subsection{Action Space Representation}

For each action dimension $i \in \{1,\ldots,d\}$, we maintain a fixed full grid $\mathcal{B}^i = \{b_1^i, \ldots, b_N^i\}$ of $N$ uniformly-spaced values covering $[a_{\min}^i, a_{\max}^i]$, and a binary active mask $m^i \in \{0,1\}^N$. Only bins $j$ with $m^i_j = 1$ participate in action selection and in the Q-value maximisation of the TD target. The active set for dimension $i$ is $\mathcal{A}^i_{\mathrm{active}} = \{b_j^i : m^i_j = 1\}$. We use $N = 9$ and initialise by activating $n_0 = 3$ bins per dimension, namely the two boundary bins and the centre bin, $j \in \{1, 5, 9\}$ (the general rule is $j_k = \mathrm{round}((k-1)(N-1)/(n_0-1))$). These three bins are \emph{anchors}: they are never pruned.

For each bin, SGQN keeps a running mean $\bar{Q}^i_j$ of the Q-value of that bin over the states on which it was evaluated since the bin was last activated (reset to zero on activation), the number of times the policy selected the bin since activation, and the number of pruning events the bin has lived through (its \emph{age}).

\subsection{Action Selection}

With probability $\varepsilon$, each dimension independently selects an active bin uniformly at random. Otherwise it selects $\argmax_{j: m^i_j = 1} Q_i(s, b^i_j)$. The exploration rate starts at $\varepsilon = 0.1$ and is multiplied by $0.995$ after every episode down to $0.01$. An earlier version of SGQN used a temperature-annealed softmax over active bins instead. We replaced it by $\varepsilon$-greedy selection, which gave more stable learning curves on tasks such as Fish Swim and Hopper Hop in our runs, and we keep it as an ablation (Section~\ref{sec:ablation}).

\subsection{Selective Growth}
\label{sec:growth}

Every $\Delta_g$ episodes after a warm-up of $T_\mathrm{warm}$ episodes, SGQN grows each dimension. For dimension $i$, the set of high-value bins is
\begin{equation}
\mathcal{H}^i = \bigl\{j : m^i_j = 1,\; \bar{Q}^i_j > \tau_Q \cdot \max_{j': m^i_{j'}=1} \bar{Q}^i_{j'}\bigr\},
\end{equation}
where $\tau_Q \in (0,1)$ is the growth threshold. For each $j \in \mathcal{H}^i$, SGQN activates the inactive immediate neighbours $j \pm 1$ that lie in the grid and are not in their regrow cooldown (a bin pruned less than $c$ pruning events ago is held back):
\begin{equation}
m^i_{j'} \leftarrow 1 \quad \forall j' \in \{j-1, j+1\} \cap \{1,\ldots,N\} \text{ with } m^i_{j'} = 0.
\end{equation}
A newly activated bin starts from a warm start. Its output weights, for both the online and the target network, are interpolated from the nearest active bins on either side, and the optimiser moments of those weights are reset to zero. This replaces the implicit warm start of \gqn, where new bins inherit their values from the shared trunk. The same rule is applied with all active bins treated as candidates in the \emph{uniform growth} ablation.

\subsection{Lazy Pruning}
\label{sec:pruning}

Every $\Delta_p$ episodes, SGQN considers each dimension for pruning. A bin is \emph{mature} if its age is at least $a_{\min}$. Among the mature active bins of a dimension, a bin is \emph{flagged} if its running mean Q-value lies below the $\tau_p$-quantile of the mature bins' running means. A flagged bin is pruned only if it is eligible, that is,
\begin{equation}
j \notin \text{anchors} \quad \text{and} \quad \bigl(\mathrm{visits}^i_j \ge v_{\min} \;\text{ or }\; \mathrm{age}^i_j \ge L_{\max}\bigr),
\end{equation}
so that either the low value has been confirmed by $v_{\min}$ selections since activation, or the policy has repeatedly declined the bin across $L_{\max}$ pruning events. A dimension never drops below $n_{\min}$ active bins. The quantile form of the threshold keeps pruning workable when converged Q-values sit in a narrow band, in which a threshold relative to the maximum would either remove everything or nothing.

\subsection{Training}
SGQN uses the same off-policy training infrastructure as \decqn\ and \gqn. A prioritised experience replay buffer \citep{schaul2016prioritized} samples transitions with probability proportional to their TD error magnitude, with a constant importance-sampling exponent $\beta = 0.4$ (full details in Appendix~\ref{app:hyperparams}). SGQN performs one gradient update per environment step. The Q-network uses the decoupled architecture of \decqn: a shared trunk with layer normalisation \citep{ba2016layer} feeds per-dimension Q-value heads, each outputting $N$ values, one per bin, and SGQN uses twin critics of this form. Only the outputs of active bins are used for selection and for the maximisation in the target. $n$-step returns with $n=3$ provide temporal smoothing of the TD target, and gradient clipping with norm 40 stabilises training. The full procedure is given in Algorithm~\ref{alg:sgqn}.

\subsection{Computational Complexity}

Like \gqn\ and \decqn, SGQN maintains $O(d \times N)$ Q-values in the network output layer. The growth and pruning operations scan the mask in $O(d \times N)$ time per check, which is negligible relative to the neural network forward pass.

\begin{algorithm}[H]
\caption{Selective Growing Q-Networks (SGQN)}
\label{alg:sgqn}
\begin{algorithmic}[1]
\Require Full grid $\mathcal{B}^i$ per dimension $i$; initial bins $n_0$; intervals $\Delta_g, \Delta_p$; thresholds $\tau_Q, \tau_p$; exploration $\varepsilon$ with decay
\State Initialise masks $m^i$ with $n_0$ anchor bins per dimension; initialise twin critics, target networks, replay buffer $\mathcal{D}$
\For{episode $= 1, 2, \ldots$}
    \State Collect an episode with $\varepsilon$-greedy selection over the active bins; decay $\varepsilon$
    \State Store transitions in $\mathcal{D}$; update running Q-means $\bar{Q}^i_j$ and selection counts
    \For{each training step}
        \State Sample a batch with prioritised replay; compute the decoupled TD loss over active bins
        \State Update the critics by gradient descent; update the target networks
    \EndFor
    \If{episode $> T_\mathrm{warm}$ and episode $\bmod \Delta_g = 0$}
        \State \textbf{Grow:} activate neighbours of high-Q bins and warm-start them (Section~\ref{sec:growth})
    \EndIf
    \If{episode $\bmod \Delta_p = 0$}
        \State \textbf{Prune:} remove flagged, eligible bins (Section~\ref{sec:pruning})
    \EndIf
\EndFor
\end{algorithmic}
\end{algorithm}

\section{Experiments}
\label{sec:experiments}

We evaluate SGQN on eight tasks from the DeepMind Control Suite \citep{tassa2018deepmind} and compare it against five baselines. Every run uses 1M environment steps, and we target 7 random seeds per method and task. We report the mean episode return and the standard deviation across seeds. A few baseline runs produced no data and are excluded; the affected table entries are marked and listed with Table~\ref{tab:final_performance}.

\subsection{Experimental Setup}

\paragraph{Tasks.} We select tasks covering a range of action dimensionalities and control difficulties: \textbf{Reacher Easy} and \textbf{Reacher Hard} (2 action dimensions), \textbf{Hopper Hop} (4), \textbf{Fish Swim} (5), \textbf{Cheetah Run}, \textbf{Walker Walk} and \textbf{Walker Run} (6), and \textbf{Quadruped Walk} (12). Hopper Hop and Walker Run are hard exploration and locomotion tasks on which several baselines fail.

\paragraph{Baselines.}
\begin{itemize}
\item \textbf{Bang-Bang PPO and MPO}: Bernoulli policies \citep{seyde2021bangbang} trained with Proximal Policy Optimisation (PPO) \citep{schulman2017proximal} and Maximum a Posteriori Policy Optimisation (MPO) \citep{abdolmaleki2018maximum}, using binary $\{-1, +1\}$ discretisation.
\item \textbf{\decqn}: Fixed 3-bin (bang-off-bang) decoupled Q-learning \citep{seyde2022solving}, the setting of that paper for state-based tasks.
\item \textbf{\gqn}: Uniform adaptive growth from 2 to 9 bins \citep{seyde2024growing}.
\item \textbf{Coarse-to-Fine Q-Networks (CQN)}: Hierarchical coarse-to-fine Q-learning, 3 levels $\times$ 5 bins \citep{seo2024continuous}, with the configuration of the official implementation, trained for 1M environment frames (500k agent steps with an action repeat of 2). CQN learns from pixels (84$\times$84 images, 3 stacked frames), whereas all other methods observe the proprioceptive state, so its returns are not a like-for-like comparison with theirs. The CQN paper also reports results for state-based tasks, which we did not run, so this comparison contains no state-based CQN.
\item \textbf{SGQN without pruning (earlier implementation)}: an earlier version of our method without pruning, with different default hyperparameters and softmax selection. It is not the ablation of Section~\ref{sec:ablation}, which changes one component of the current method at a time.
\end{itemize}

\paragraph{Hyperparameters.} Each baseline uses the hyperparameters stated in its original publication. Where the publication does not state a value, or no official configuration exists, the baseline uses the default of our code, which Table~\ref{tab:components} marks with C. SGQN uses layer normalisation, ELU activations, two hidden layers of 512 units, Adam optimiser with learning rate $10^{-4}$, discount $\gamma = 0.99$, replay buffer size $10^6$, batch size 256, 3-step returns, target network hard update every 100 gradient updates, prioritised replay with $\alpha = 0.6$ and a constant $\beta = 0.4$ (Appendix~\ref{app:hyperparams}). SGQN-specific hyperparameters are listed in Table~\ref{tab:sgqn_hyperparams}. They were set by a small manual search on the tasks and are examined in Section~\ref{sec:ablation}.

\paragraph{Differences between SGQN and the baselines.} Table~\ref{tab:components} lists the components of SGQN, \gqn, \decqn\ and CQN as implemented in our code and, where the original papers state them, as specified there. Three points matter for reading the results. First, SGQN performs one gradient update per environment step, whereas our \gqn\ and \decqn\ runs perform one per eight (a sample-to-insert ratio of 32 with batch size 256 in the code of both baselines; the \decqn\ and \gqn\ papers do not state this ratio, and we could not check the official \decqn\ code). With a target update every 100 updates, the target network of SGQN is also refreshed every 100 environment steps against every 800 for \gqn. Second, SGQN decays $\varepsilon$ from 0.1 to 0.01, whereas \gqn\ and \decqn\ keep it at 0.1, and it uses a different network and a different prioritised-replay exponent $\beta$. Third, the pieces that are specific to SGQN are the pruning rule, the value-guided growth rule, growth at fixed episodes from three bins, and the interpolated warm start of new bins. Two averaged critics, 3-step returns, the prioritised-replay exponent $\alpha$ and the optimiser settings are shared with \gqn\ and \decqn. Bang-Bang PPO and MPO are policy-gradient methods and have none of these components. None of the differences in update frequency, $\varepsilon$ schedule, $\beta$, network or warm start is ablated in this paper.

\begin{table}[t]
\centering
\small
\caption{Components of SGQN and of the Q-learning baselines. C: our implementation, P: the original paper. ``Not stated'' means the paper does not specify it. Bang-Bang PPO and MPO are policy-gradient methods and are omitted.}
\label{tab:components}
\resizebox{\textwidth}{!}{%
\begin{tabular}{p{3.1cm}p{3.6cm}p{3.6cm}p{3.6cm}p{3.6cm}}
\toprule
Component & SGQN & \gqn & \decqn & CQN \\
\midrule
Observations & State & State & State & Pixels in our runs (84$\times$84, 3 frames); the paper also reports state-based tasks \\
Two averaged critics & Yes & Yes (C); double Q-learning (P) & Yes (C, P) & Not stated \\
$n$-step, $\alpha$ & 3, 0.6 & 3, 0.6 & 3, 0.6 (C, P) & 3 (1 on Walker tasks) \\
Replay exponent $\beta$ & 0.4, constant & 0.2, constant & 0.2 (C, P) & Not applicable \\
Network & Layer-norm MLP trunk, 2$\times$512 & Residual layer-norm network, 512 & Same as \gqn\ (C); 500 hidden units (P) & Pixel encoder \\
Updates per environment step & 1 & 1/8 (C); not stated (P) & 1/8 (C); not stated (P) & 1 per 2 agent steps \\
Exploration & $\varepsilon$-greedy, 0.1 decayed to 0.01 & $\varepsilon$-greedy, constant 0.1 (C); not stated (P) & $\varepsilon$-greedy, constant 0.1 (C, P) & Gaussian noise, std 0.1 (P) \\
Bins at the start & 3 & 2, then 3, 5, 9 & 3, fixed & 3 levels $\times$ 5 bins \\
Growth trigger & Fixed episodes (300, 600) & Stagnation criterion & None & None \\
Bins that grow & Neighbours of high-Q bins (in practice all) & All, uniformly & None & None \\
Pruning & Yes & No & No & No \\
Warm start of new bins & Interpolated neighbour weights & Shared network only & None & None \\
\bottomrule
\end{tabular}}
\end{table}

\subsection{Main Results}

\begin{figure}[t]
\centering
\subfloat[Reacher Easy.\label{fig:reacher_easy}]{%
    \includegraphics[width=0.32\textwidth]{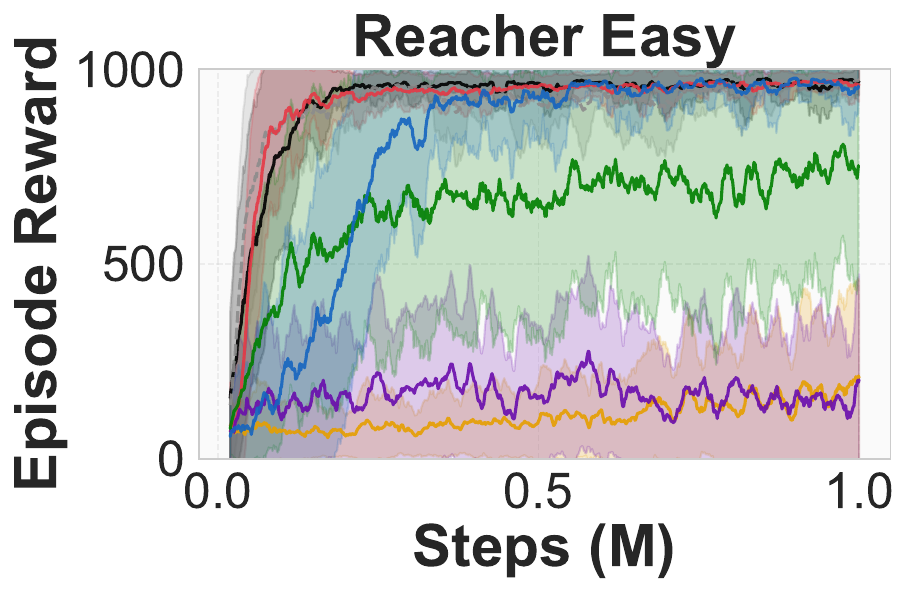}}%
\hfill%
\subfloat[Reacher Hard.\label{fig:reacher_hard}]{%
    \includegraphics[width=0.32\textwidth]{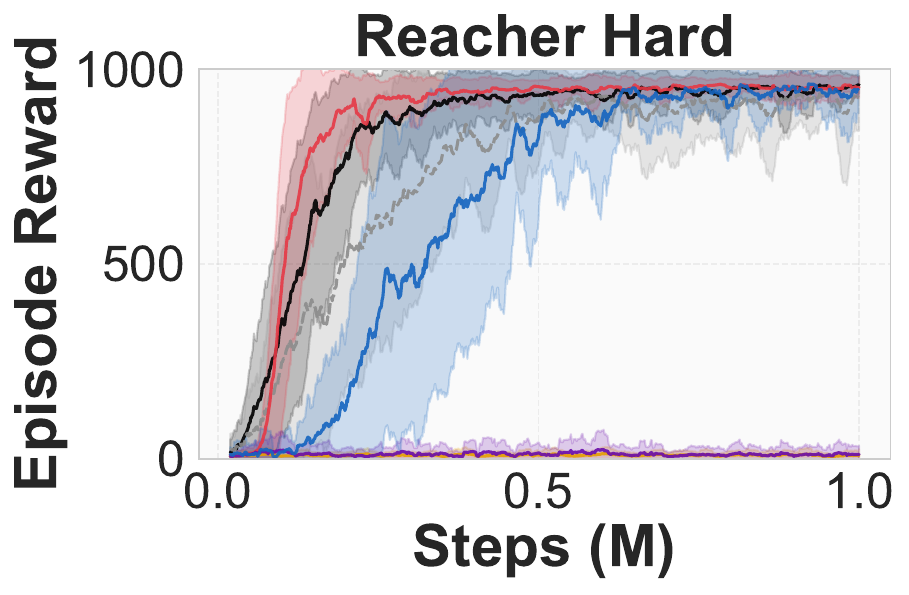}}%
\hfill%
\subfloat[Hopper Hop.\label{fig:hopper_hop}]{%
    \includegraphics[width=0.32\textwidth]{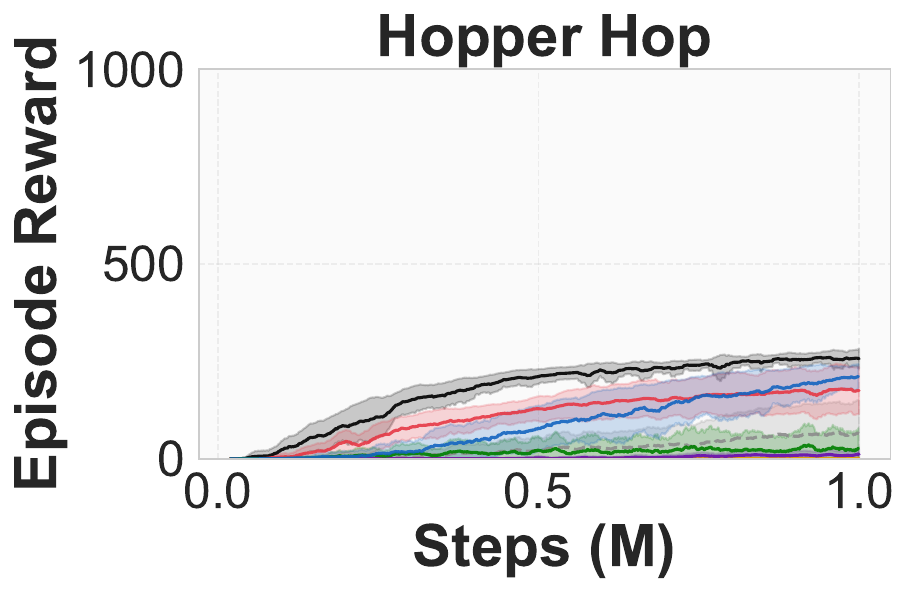}}\\[0.5ex]
\subfloat[Fish Swim.\label{fig:fish_swim}]{%
    \includegraphics[width=0.32\textwidth]{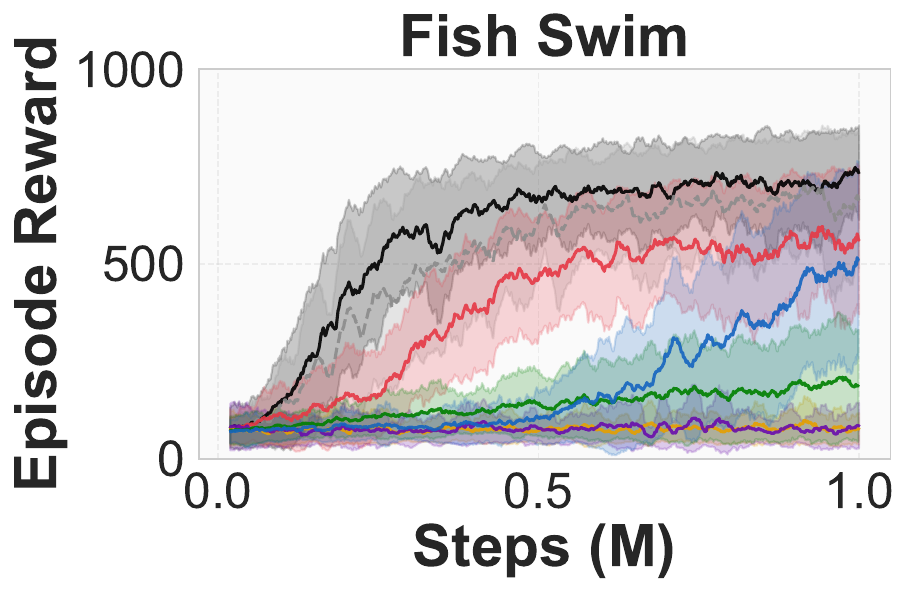}}%
\hfill%
\subfloat[Cheetah Run.\label{fig:cheetah_run}]{%
    \includegraphics[width=0.32\textwidth]{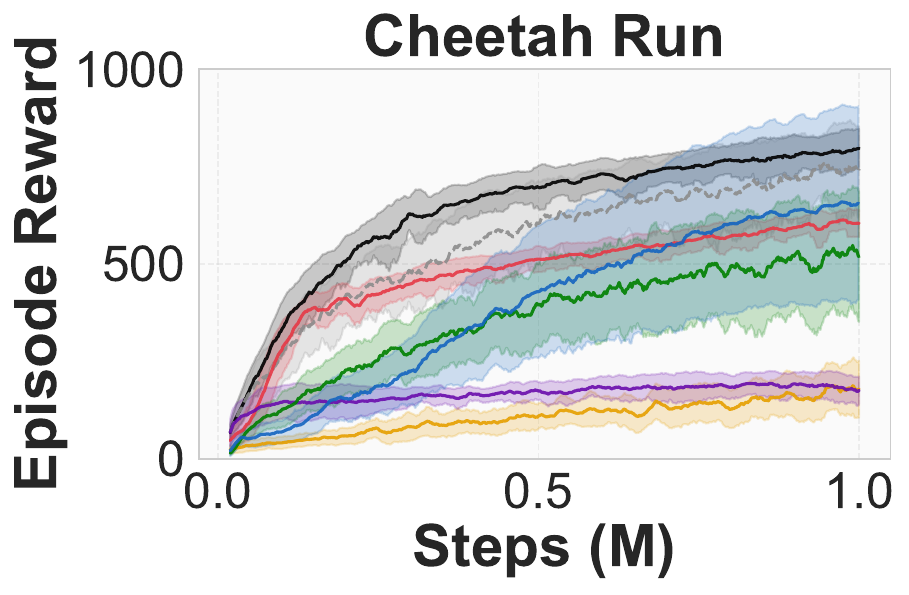}}%
\hfill%
\subfloat[Walker Walk.\label{fig:walker_walk}]{%
    \includegraphics[width=0.32\textwidth]{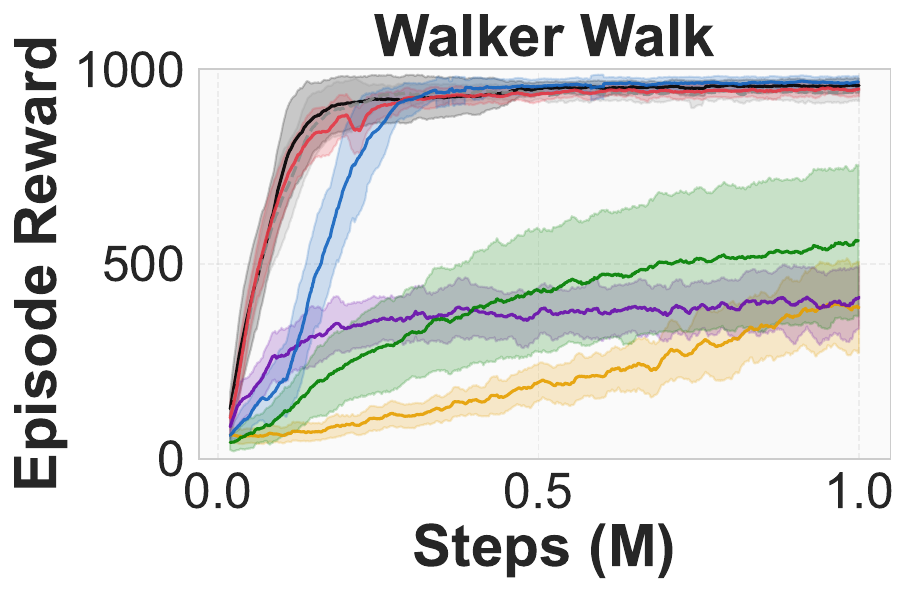}}\\[0.5ex]
\subfloat[Walker Run.\label{fig:walker_run}]{%
    \includegraphics[width=0.32\textwidth]{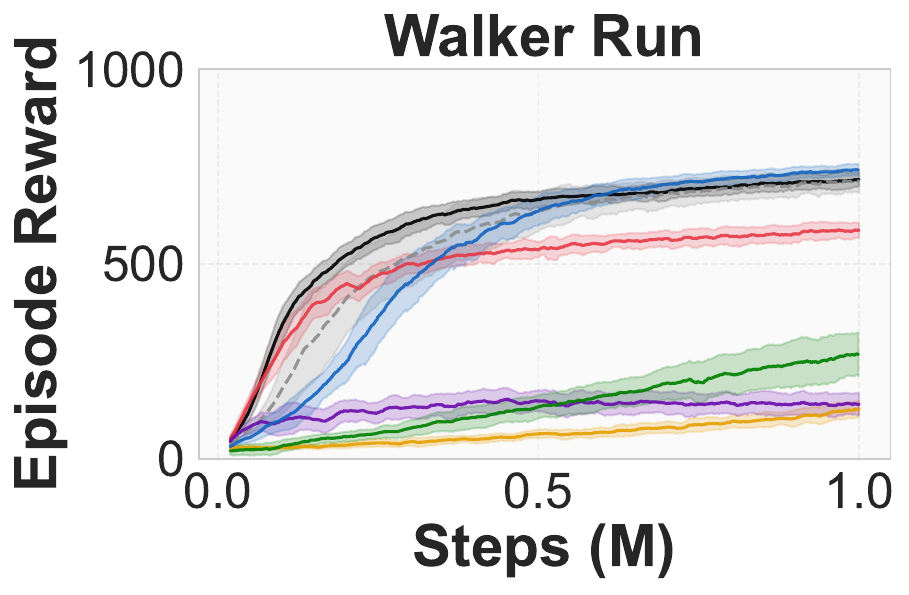}}%
\hspace{0.03\textwidth}%
\subfloat[Quadruped Walk.\label{fig:quadruped_walk}]{%
    \includegraphics[width=0.32\textwidth]{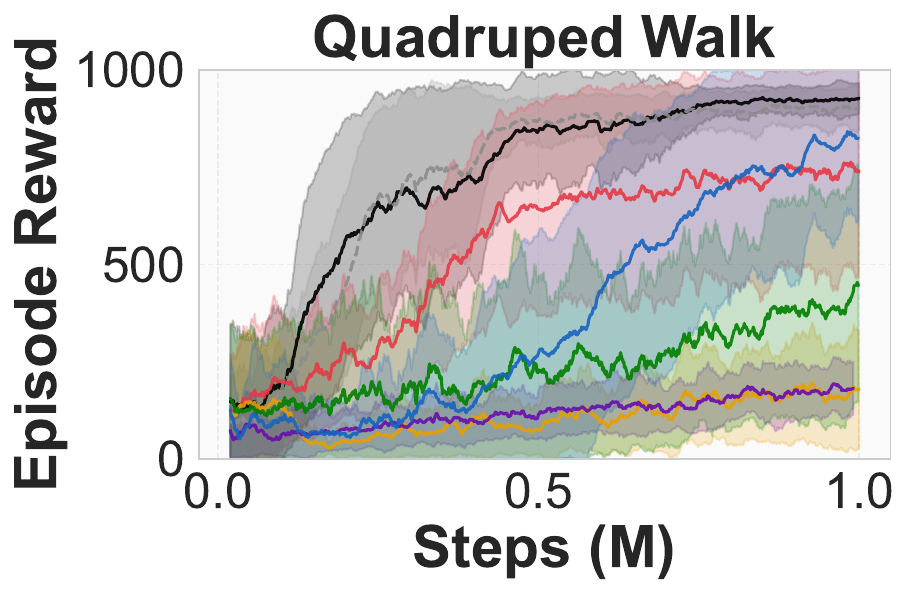}}\\[1.0ex]
\includegraphics[width=0.80\textwidth]{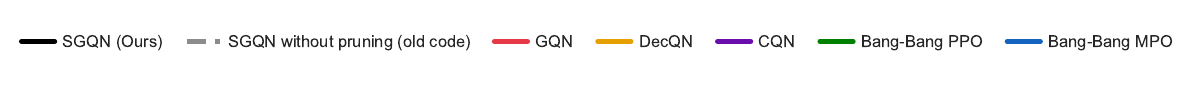}
\caption{Learning curves on eight DeepMind Control Suite tasks (mean $\pm$ standard deviation over the seeds that produced data, 7 for most entries, see Table~\ref{tab:final_performance}; 1M environment steps).}
\label{fig:learning_curves}
\end{figure}

Table~\ref{tab:final_performance} reports the mean episode return over the final 100 training episodes.

\begin{table}[t]
\centering
\caption{Mean training episode return over the final 100 episodes (mean\,$\pm$\,standard deviation over seeds). \textbf{Bold}: highest mean per task over all rows. $^\dagger$: 6 instead of 7 seeds, because the remaining run produced no data (Bang-Bang PPO: Cheetah Run and Reacher Easy; Bang-Bang MPO: Cheetah Run; CQN: Cheetah Run and Quadruped Walk; earlier SGQN version: Quadruped Walk). --: none of the seven runs produced data. CQN learns from pixels, all other methods from the state.}
\label{tab:final_performance}
\resizebox{\textwidth}{!}{%
\begin{tabular}{lcccccccc}
\toprule
Method & Cheetah Run & Fish Swim & Hopper Hop & Quadruped Walk & Reacher Easy & Reacher Hard & Walker Run & Walker Walk \\
\midrule
SGQN (Ours) & \textbf{788 $\pm$ 37} & \textbf{715 $\pm$ 17} & \textbf{257 $\pm$ 11} & \textbf{924 $\pm$ 10} & 962 $\pm$ 12 & 951 $\pm$ 16 & 713 $\pm$ 13 & 957 $\pm$ 3 \\
SGQN without pruning (old code) & 741 $\pm$ 42 & 662 $\pm$ 51 & 63 $\pm$ 76 & 901 $\pm$ 15$^\dagger$ & \textbf{967 $\pm$ 6} & 924 $\pm$ 67 & 710 $\pm$ 22 & 941 $\pm$ 5 \\
GQN & 603 $\pm$ 14 & 561 $\pm$ 19 & 174 $\pm$ 55 & 739 $\pm$ 208 & 966 $\pm$ 4 & \textbf{953 $\pm$ 4} & 585 $\pm$ 11 & 948 $\pm$ 2 \\
DecQN & 172 $\pm$ 18 & 80 $\pm$ 4 & 1 $\pm$ 0 & 168 $\pm$ 30 & 182 $\pm$ 89 & 9 $\pm$ 2 & 119 $\pm$ 8 & 389 $\pm$ 93 \\
CQN & 179 $\pm$ 10$^\dagger$ & 79 $\pm$ 7 & 10 $\pm$ 4 & 179 $\pm$ 23$^\dagger$ & 168 $\pm$ 61 & 11 $\pm$ 7 & 144 $\pm$ 18 & 413 $\pm$ 49 \\
Bang-Bang PPO & 515 $\pm$ 97$^\dagger$ & 193 $\pm$ 123 & 27 $\pm$ 31 & 396 $\pm$ 197 & 754 $\pm$ 46$^\dagger$ & -- & 258 $\pm$ 49 & 548 $\pm$ 186 \\
Bang-Bang MPO & 648 $\pm$ 244$^\dagger$ & 485 $\pm$ 136 & 203 $\pm$ 30 & 812 $\pm$ 199 & 959 $\pm$ 17 & 943 $\pm$ 17 & \textbf{738 $\pm$ 5} & \textbf{967 $\pm$ 2} \\
\bottomrule
\end{tabular}
}
\end{table}

\paragraph{Where SGQN is best.} SGQN has the highest mean return on four tasks. On Cheetah Run it reaches 788, against 741 for the earlier SGQN version, 648 for Bang-Bang MPO and 603 for \gqn. On Fish Swim it reaches 715, against 662 for the earlier version and 561 for \gqn. On Hopper Hop it reaches 257, against 203 for Bang-Bang MPO and 174 for \gqn, and the earlier version learns little (63). On Quadruped Walk, the task with 12 action dimensions, it reaches 924, against 901 for the earlier version, 812 for Bang-Bang MPO and 739 for \gqn. The margin to the earlier SGQN version is about one standard deviation of the seeds on Cheetah Run and Fish Swim, and the margins to all other methods are larger. On Hopper Hop and Quadruped Walk SGQN also has a much smaller spread over seeds (standard deviations of 11 and 10) than \gqn\ and Bang-Bang MPO, which have 55 and 30 on Hopper Hop and 208 and 199 on Quadruped Walk.

\paragraph{Reacher tasks.} The 2-dimensional Reacher tasks do not separate the strongest methods. On Reacher Easy, SGQN (962), \gqn\ (966), the earlier SGQN version (967) and Bang-Bang MPO (959) are within 1\% of each other, and on Reacher Hard SGQN (951), \gqn\ (953) and Bang-Bang MPO (943) are as well. All of these differences are within about one standard deviation of the seeds, so we do not treat any of these methods as better on these tasks.

\paragraph{Where SGQN is not best.} Bang-Bang MPO has the highest mean on both Walker tasks: 738 against 713 for SGQN on Walker Run (3.4\% lower), and 967 against 957 on Walker Walk (1.0\% lower). Both gaps are larger than the standard deviations across seeds of either method (13 and 5 on Walker Run, 3 and 2 on Walker Walk), so Bang-Bang MPO is better on these two tasks in our runs. SGQN is ahead of \gqn\ on both (585 and 948), and far ahead of Bang-Bang PPO (258 and 548).

\paragraph{Weak baselines.} \decqn\ and CQN stay below every other method on every task, mostly by a wide margin: their best results are 389 and 413 on Walker Walk, and they reach 1 and 10 on Hopper Hop and 9 and 11 on Reacher Hard. We ran both with the settings of their publications and did not tune them, and we did not investigate the gap further. The comparison with them therefore says little about SGQN beyond the fact that it does not share their failures, and the CQN numbers in particular are not like-for-like because CQN learns from pixels.

\paragraph{What the comparison shows.} Over the eight tasks SGQN is ahead of \gqn\ on six (by 185 on Cheetah Run and 154 on Fish Swim, and by 128 on Walker Run), and behind it by 4 and 2 points on the two Reacher tasks, which is within noise. This advantage should not be attributed to the adaptive grid, because the ablations of Section~\ref{sec:ablation} show that value-guided growth and pruning do not change the return. SGQN and \gqn\ differ in other respects as well (Table~\ref{tab:components}): SGQN performs one gradient update per environment step against one per eight for \gqn, decays $\varepsilon$ from 0.1 to 0.01 where \gqn\ keeps it at 0.1, uses a different network and replay exponent $\beta$, starts from three bins instead of two and grows on a fixed schedule instead of the stagnation criterion, and warm-starts new bins from their neighbours. Both use two averaged critics. We did not run \gqn\ with the update frequency or other settings of SGQN, so we cannot say which of these differences accounts for the advantage. The larger number of updates per environment step is a candidate that we have not tested.

\subsection{Where Does SGQN Allocate Resolution?}
\label{sec:allocation}

The motivation of SGQN is that resolution should be concentrated where the value function indicates it matters. We test this directly on the 56 runs (8 tasks, 7 seeds) of the main experiment. Every 25 episodes the run records, for each dimension and bin, whether it is active, its running mean Q-value, and how often it was selected, so that the grid can be compared to the Q-values over the whole training run. Throughout, ``Q-value of a bin'' is the running mean described in Section~\ref{sec:method}.

\begin{figure}[t]
\centering
\includegraphics[width=\textwidth]{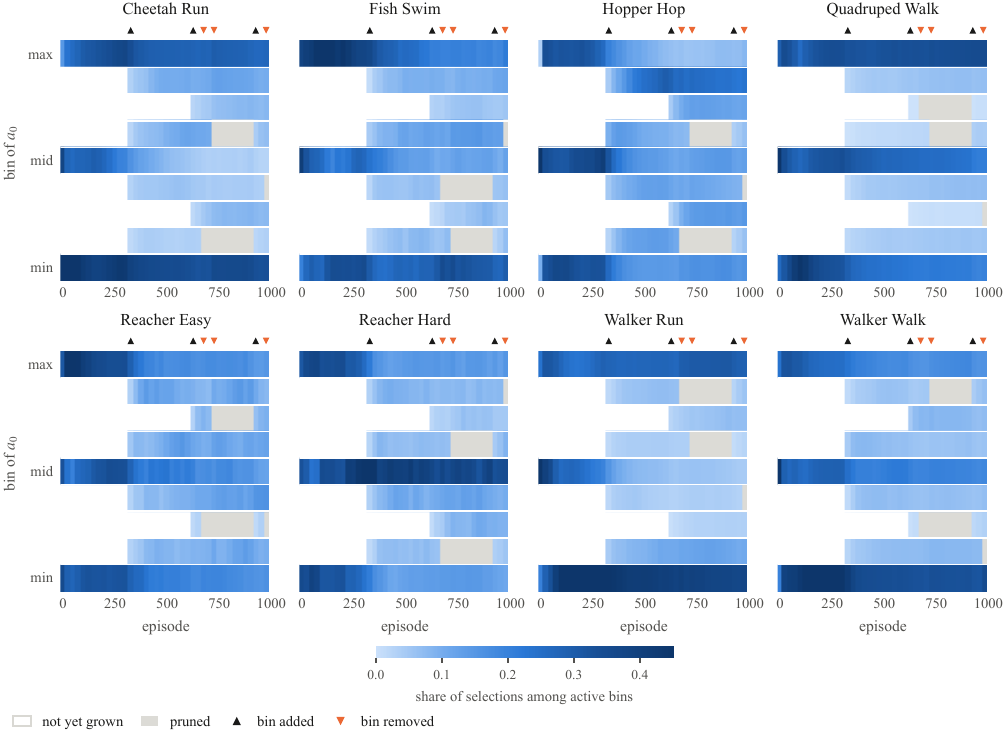}
\caption{Active action bins of the first action dimension over training, for one seed per task. Rows are the nine bins (action value increasing upwards), and cell colour is the share of the selections of that dimension that went to the bin in each interval. White cells are bins that have not yet been grown into the grid, gray cells are bins that were pruned, and triangles mark bins that were added (upward) or removed (downward) at that checkpoint.}
\label{fig:bin_lanes}
\end{figure}

\begin{figure}[t]
\centering
\includegraphics[width=\textwidth]{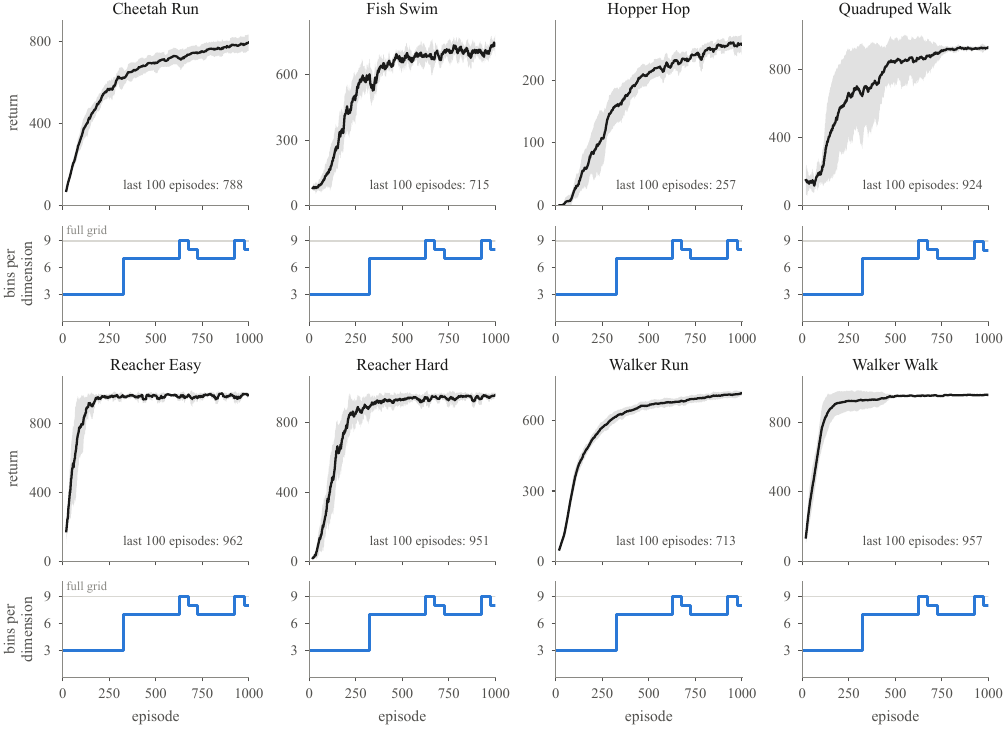}
\caption{Return (top of each block, mean $\pm$ standard deviation over seeds, window of 20 episodes) and number of active bins per dimension (bottom of each block, mean over seeds and dimensions) over training for the eight tasks. The step profile of the bin count is the same on every task.}
\label{fig:resolution}
\end{figure}

\paragraph{Which bins are active.} Figure~\ref{fig:bin_lanes} shows the grid of one dimension of one seed over training. Every task follows the same schedule: three anchor bins until the first growth step at episode 300, seven active bins after it, nine after the growth step at episode 600, and then a cycle of pruning (one bin removed at each of episodes 650 and 700), regrowth at episode 900 and further pruning at episode 950. The schedule is identical across seeds and tasks (Figure~\ref{fig:resolution}), because it is determined by $\Delta_g$, $\Delta_p$ and the regrow cooldown rather than by task-specific values. Averaged over the last 300 episodes (one growth interval), 7.5 of the 9 bins per dimension are active on every task (Table~\ref{tab:final_resolution}).

\paragraph{Growth does not select.} The growth rule of Section~\ref{sec:growth} activates neighbours of bins whose running mean Q-value exceeds $0.85$ times the best bin of the dimension. In our runs it activated every candidate bin in every dimension: the grid always grows from three to seven to nine bins, as it would under uniform growth. The reason is visible in the Q-values themselves. At the last checkpoint, the range of Q over the nine bins of a dimension is a median of 0.5\% to 2.9\% of its mean on the eight tasks (Table~\ref{tab:final_resolution}). With such a narrow spread, almost every bin lies above $0.85$ times the maximum. The alignment of the active bins with the Q-values, measured as the mean Q of the active bins divided by the mean Q over the whole grid, is 1.000 on every task and 1.001 on Quadruped Walk. In other words, in these runs the Q-values of the bins that SGQN keeps cannot be distinguished from those of the bins it does not, and we found no evidence that growth concentrates resolution in high-value regions. A higher threshold $\tau_Q$ makes growth stricter, and we vary it up to $0.95$ in Section~\ref{sec:ablation}, where it also leaves the return essentially unchanged.

\begin{table}[t]
\centering
\small
\caption{Action-space resolution of SGQN at the end of training. \emph{Active bins} is the number of active bins per dimension averaged over the last 300 episodes, which is one full growth interval (the count rises when growth fires and falls at each pruning event, so the last checkpoint alone depends on where in that cycle training stopped). The grid has $N=9$ bins per dimension. \emph{Alignment} is the mean Q of the active bins divided by the mean Q of the full grid at the last checkpoint, and \emph{Q spread} is the range of Q over the nine bins of a dimension as a share of its mean (median over dimensions). Return is the mean over the final 100 training episodes. All values are means over 7 seeds.}
\label{tab:final_resolution}
\begin{tabular}{lccccc}
\toprule
Task & Action dim. & Active bins & Alignment & Q spread & Return \\
\midrule
Cheetah Run & 6 & 7.5 / 9 & 1.000 & 0.8\% & 788 \\
Fish Swim & 5 & 7.5 / 9 & 1.000 & 1.0\% & 715 \\
Hopper Hop & 4 & 7.5 / 9 & 1.000 & 2.9\% & 257 \\
Quadruped Walk & 12 & 7.5 / 9 & 1.001 & 2.6\% & 924 \\
Reacher Easy & 2 & 7.5 / 9 & 1.000 & 0.5\% & 962 \\
Reacher Hard & 2 & 7.5 / 9 & 1.000 & 0.6\% & 951 \\
Walker Run & 6 & 7.5 / 9 & 1.000 & 1.1\% & 713 \\
Walker Walk & 6 & 7.5 / 9 & 1.000 & 1.0\% & 957 \\
\bottomrule
\end{tabular}
\end{table}

\begin{figure}[t]
\centering
\includegraphics[width=\textwidth]{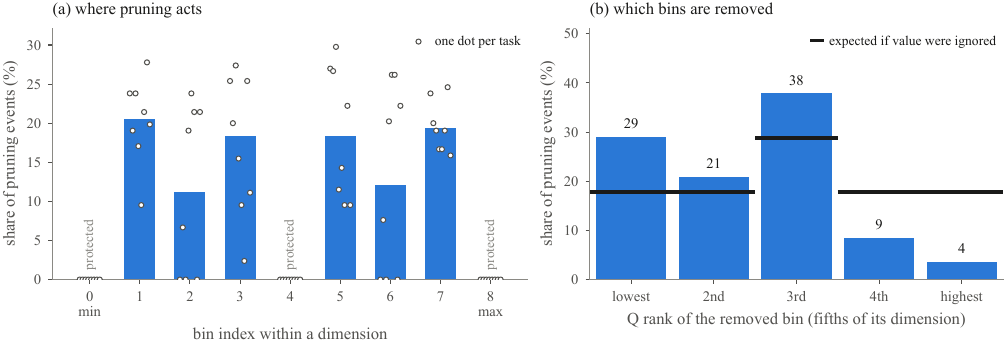}
\caption{Which bins are pruned. (a) Share of all pruning events that removed each bin position of the grid, with the share on each individual task as dots. Anchor bins are protected and are never removed. (b) Distribution over the fifths of the Q-ranking (lowest fifth on the left) of the bins that were removed, with the share expected under random removal, computed per event from the number of candidates, as ticks.}
\label{fig:specialisation}
\end{figure}

\paragraph{Pruning is weakly value-guided.} Across the 56 runs there were 903 pruning events. None removed an anchor bin, as intended. For each event, we ranked the bins that were candidates at that moment by their running mean Q-value and recorded the fifth of the ranking that the removed bin fell into (Figure~\ref{fig:specialisation}b). The removed bin was in the lowest fifth in 29\% of events, against 17.8\% expected if a candidate were removed at random given the candidate counts of each event, and in the lowest two fifths in 50\% of events, against 36\% expected. Pruning is therefore biased towards low-Q bins, but the bias is moderate, and about half of the removed bins are not in the lower two fifths. Because the running mean is taken since activation, it is also biased by the age of a bin, which may contribute to which bins are removed. We have not isolated this effect, and we treat the explanation as a hypothesis. Figure~\ref{fig:specialisation}a shows which bin positions are removed: removals fall on the non-anchor bins, and the distribution over positions is similar across tasks.

\paragraph{Summary.} The measurements do not support the statement that SGQN \emph{focuses representational capacity where it matters most}, if this is read as resolution following the value function. They support a narrower statement: SGQN reaches strong returns on all eight tasks with a grid that is on average smaller than the full grid and that, in these runs, is determined mainly by the growth and pruning schedule, with a weak preference for removing low-Q bins. Whether the performance comes from the adaptive grid itself, from starting coarse, or from $\varepsilon$-greedy selection is what the ablations below separate.

\subsection{Ablations and Hyperparameter Sensitivity}
\label{sec:ablation}

\paragraph{Component ablation.} We change one design choice at a time and keep the rest of the method fixed, on Quadruped Walk, Hopper Hop, Walker Run and Fish Swim with 5 seeds each: (i) no pruning, (ii) uniform instead of value-guided growth (all active bins attract their neighbours), (iii) the full grid of 9 bins from the start, which removes growth and pruning altogether, and (iv) the earlier annealed softmax selection instead of $\varepsilon$-greedy. Since growth already behaves like uniform growth in the main runs (Section~\ref{sec:allocation}), variant (ii) is expected to match the full method, and we report it to confirm that.

\paragraph{Hyperparameter sensitivity.} We vary one hyperparameter at a time around the defaults, on Hopper Hop and Walker Run with 3 seeds: the pruning threshold $\tau_p \in \{0.25, 0.5, 0.75\}$, the growth threshold $\tau_Q \in \{0.7, 0.85, 0.95\}$, the pruning interval $\Delta_p \in \{25, 50, 100\}$, the growth interval $\Delta_g \in \{150, 300, 600\}$, the exploration decay in $\{0.99, 0.995, 0.999\}$, and the number of initial bins $n_0 \in \{3, 5, 7\}$. The defaults were chosen by a small manual search on the main tasks before these experiments, and the study shows how much of the performance depends on that choice.

\begin{table}[t]
\centering
\small
\caption{Component ablation. Each row changes one design choice and leaves the rest of the full method unchanged. Values are the mean return over the final 100 training episodes, with the standard deviation across seeds in parentheses. All variants use the same 5 seeds.}
\label{tab:ablation}
\begin{tabular}{lcccc}
\toprule
Variant & Fish Swim & Hopper Hop & Quadruped Walk & Walker Run \\
\midrule
Full method & 718 (13) & 254 (12) & 927 (10) & 713 (15) \\
No pruning & 726 (12) & 255 (12) & 928 (8) & 710 (16) \\
Uniform instead of value-guided growth & 718 (13) & 254 (12) & 926 (11) & 713 (15) \\
Start with all 9 bins (no growth) & 722 (14) & 252 (32) & 818 (132) & 684 (49) \\
Annealed softmax instead of $\varepsilon$-greedy & 508 (19) & 0 (0) & 250 (44) & 547 (21) \\
\bottomrule
\end{tabular}
\end{table}

\begin{figure}[t]
\centering
\includegraphics[width=\textwidth]{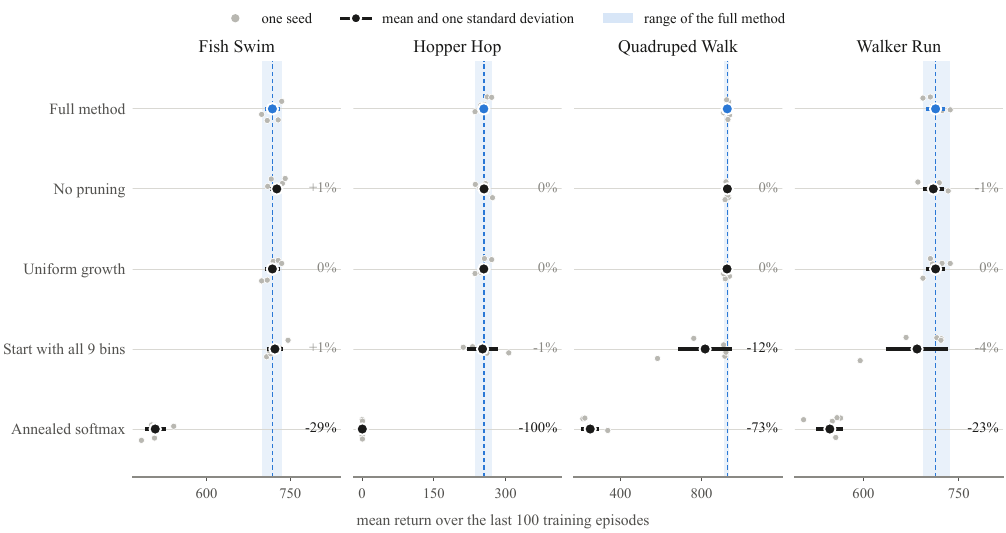}
\caption{Component ablation on four tasks with 5 seeds each. Gray dots are single seeds, dark markers are means with one standard deviation, and the shaded band is the range of the full method over the same seeds. The percentage at the right of each row is the mean relative to the full method. Uniform growth and no pruning stay inside the band, starting with all 9 bins produces individual seeds far below it, and annealed softmax is far below it on every task.}
\label{fig:ablation}
\end{figure}

\paragraph{Uniform growth is indistinguishable from value-guided growth.} The returns of the two variants agree to the rounded value in 19 of 20 runs, and the remaining run (Quadruped Walk, seed 21) differs by 7 points (917 against 924). This follows from Section~\ref{sec:allocation}: with our thresholds the growth rule already activates every candidate bin, so replacing it by uniform growth leaves the run unchanged. It does not show that value-guided growth is harmless in general, only that in this setting it does not act differently from uniform growth.

\paragraph{Pruning does not change the return.} Without pruning, the mean return differs from the full method by between $-0.4\%$ and $+1.1\%$ on the four tasks, within one standard deviation across seeds in every case. Pruning removes about 1.5 of the 9 bins per dimension at the end of training (Table~\ref{tab:final_resolution}) without a measurable effect on the return, so its benefit in our experiments is a smaller grid, not a better policy.

\paragraph{Starting coarse makes training more reliable.} Starting with all 9 bins gives similar means on Fish Swim (722) and Hopper Hop (252), but a much larger spread on Hopper Hop (standard deviation 32 against 12), and lower means on Quadruped Walk (818 against 927) and Walker Run (684 against 713). The lower means come from single seeds that fail: two on Quadruped Walk (761 and 582) and one on Walker Run (595), while the other seeds are close to the full method. Growing from three anchor bins therefore protects against occasional failures. Because uniform growth gives the same results as the full method, this benefit comes from starting coarse and growing, not from the choice of which bins are grown.

\paragraph{$\varepsilon$-greedy selection is the largest effect.} Annealed softmax lowers the return by 29\%, 100\%, 73\% and 23\% on Fish Swim, Hopper Hop, Quadruped Walk and Walker Run, and on Hopper Hop it does not learn (all five seeds stay below 2). The earlier implementation, which used softmax together with other defaults, reached 710 on Walker Run, so softmax is not inherently unable to learn. We did not tune its temperature schedule separately, so the ablation shows that $\varepsilon$-greedy works with our defaults and softmax does not, not that softmax cannot work.

\begin{table}[t]
\centering
\small
\caption{Hyperparameter sensitivity. Each hyperparameter is varied alone with the rest at their defaults, which are marked $\ast$. Values are the mean return over the final 100 training episodes, with the standard deviation across seeds in parentheses.}
\label{tab:sensitivity}
\begin{tabular}{lccc}
\toprule
Hyperparameter & Value & Hopper Hop & Walker Run \\
\midrule
prune threshold & 0.25 & 257 (15) & 710 (7) \\
 & 0.5 $\ast$ & 256 (15) & 712 (8) \\
 & 0.75 & 257 (13) & 713 (11) \\
\addlinespace
growth threshold & 0.7 & 256 (15) & 712 (8) \\
 & 0.85 $\ast$ & 256 (15) & 712 (8) \\
 & 0.95 & 254 (14) & 712 (8) \\
\addlinespace
prune interval & 25 & 254 (15) & 711 (8) \\
 & 50 $\ast$ & 256 (15) & 712 (8) \\
 & 100 & 257 (12) & 712 (8) \\
\addlinespace
growth interval & 150 & 250 (8) & 724 (13) \\
 & 300 $\ast$ & 256 (15) & 712 (8) \\
 & 600 & 270 (6) & 727 (7) \\
\addlinespace
epsilon decay & 0.99 & 257 (22) & 711 (23) \\
 & 0.995 $\ast$ & 256 (15) & 712 (8) \\
 & 0.999 & 233 (49) & 691 (5) \\
\addlinespace
initial bins & 3 $\ast$ & 256 (15) & 712 (8) \\
 & 5 & 226 (7) & 729 (6) \\
 & 7 & 228 (11) & 730 (5) \\
\addlinespace
\bottomrule
\end{tabular}
\end{table}

\begin{figure}[t]
\centering
\includegraphics[width=\textwidth]{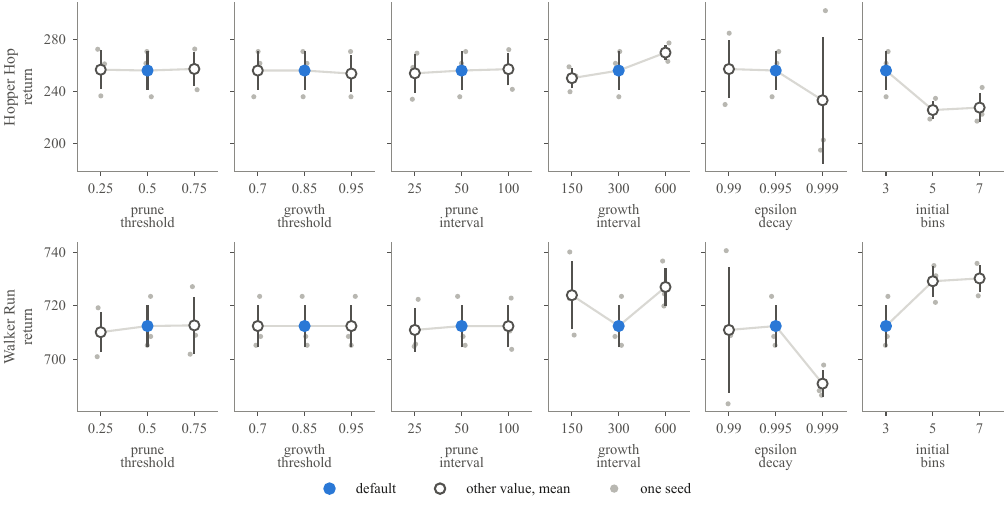}
\caption{Hyperparameter sensitivity on Hopper Hop and Walker Run with 3 seeds. Each panel varies one hyperparameter with the others at their defaults. The blue marker is the default, open markers are other values, gray dots are single seeds and bars are one standard deviation. The values on the horizontal axis are evenly spaced in increasing order.}
\label{fig:sensitivity}
\end{figure}

\paragraph{The growth and pruning hyperparameters hardly matter.} Varying $\tau_p$, $\tau_Q$ or $\Delta_p$ over the tested values changes the mean return by at most 3 points on either task, about 1\% on Hopper Hop and 0.4\% on Walker Run. A growth threshold of 0.7 gives exactly the same returns as 0.85, which again shows that growth does not discriminate between bins, and at 0.95 the return changes by at most 2 points. The defaults are therefore not finely tuned in these respects.

\paragraph{Exploration, growth interval and initial resolution matter more.} An exploration decay of 0.999 lowers the return on both tasks (233 on Hopper Hop and 691 on Walker Run, against 256 and 712), because exploration stays high for longer, while a faster decay of 0.99 matches the default. A growth interval of 600 gives higher returns on both tasks (270 and 727), and 150 gives a lower return on Hopper Hop (250) and a higher one on Walker Run (724). With 3 seeds we do not regard these differences as significant, but they indicate that the default $\Delta_g = 300$ is not optimal. The number of initial bins acts in opposite directions: 5 or 7 initial bins lower the return on Hopper Hop (226 and 228 against 256) and raise it on Walker Run (729 and 730 against 712). The best starting resolution is therefore task dependent in our runs, and 3 is not uniformly best.

\paragraph{Summary of the ablations.} The results are consistent. Among the components we varied, the return of SGQN depends on the coarse start with growth and on $\varepsilon$-greedy selection, and not on value-guided growth or on pruning. These ablations are within SGQN. They do not show why SGQN outperforms the baselines, which share a coarse start and $\varepsilon$-greedy selection. A fixed coarse grid of three bins without growth is not part of our ablation, so we cannot separate the benefit of growth from the benefit of starting coarse.

\section{Discussion}
\label{sec:discussion}

\paragraph{What the adaptive grid does.} The original motivation was that a decoupled Q-function has a peaked landscape around the optimal action, so that selective growth should concentrate new bins near the peak. Our analysis does not show this. At the thresholds we use, the Q-values of the nine bins of a dimension differ by about one percent, which is too little for the growth rule to discriminate between them, and growth acts as uniform growth. Within SGQN, the ablation of Section~\ref{sec:ablation} shows that starting coarse and $\varepsilon$-greedy selection matter, while value-guided growth and pruning do not change the return. This does not explain the advantage over the baselines, because \gqn\ and \decqn\ also start coarse and use $\varepsilon$-greedy selection, and because SGQN differs from them in update frequency and other settings (Table~\ref{tab:components}). The warm start of new bins, the update frequency and the network were not ablated, and a fixed coarse grid without growth is missing, so we cannot say whether growth itself contributes beyond starting coarse. A possible reason for the flat Q-values is that the Q-value of a bin is evaluated for a state-action pair that is averaged over states, so that differences between bins are small compared to differences between states.

\paragraph{How SGQN relates to the baselines.} SGQN has the highest mean return on four of eight tasks and is within 1.1\% of the best method on three more, but it is not uniformly best: Bang-Bang MPO is better on both Walker tasks, and \gqn\ is nominally ahead on the two Reacher tasks by margins within the seed noise. Because the adaptive grid is not what drives the result, the comparison supports the claim that SGQN as configured is competitive with the strongest baselines here, and not the claim that adaptive discretisation is better than uniform growth. Because SGQN also differs from the baselines in update frequency and other settings, it does not show which of its components is responsible.

\paragraph{The role of lazy pruning.} Pruning removes bins with lower Q-values than a random choice would, but the preference is moderate (Section~\ref{sec:allocation}). The eligibility conditions mean that a bin which the policy repeatedly declines becomes prunable after $L_{\max}$ pruning events even without many selections, so that flagged bins are not held back for long. Removing pruning changes the return by at most 1.1\% (Section~\ref{sec:ablation}), so pruning reduces the size of the grid without a measurable effect on the return.

\paragraph{Why $\varepsilon$-greedy instead of softmax.} An earlier version of SGQN used a temperature-annealed softmax over the Q-values of the active bins. With Q-values that differ by about one percent across bins, a softmax over them is close to uniform unless the temperature is very small, and the schedule of the temperature then determines the behaviour. $\varepsilon$-greedy has a single, interpretable exploration parameter, and in the ablation of Section~\ref{sec:ablation} the softmax variant reaches 23\% to 100\% lower returns and does not learn on Hopper Hop. The earlier implementation, which used softmax with other defaults, did learn on Walker Run, so the softmax schedule and the other hyperparameters interact, and we did not tune softmax separately.

\paragraph{Limitations.} SGQN introduces hyperparameters compared to \gqn\ ($\tau_Q$, $\tau_p$, $\Delta_g$, $\Delta_p$, $v_{\min}$, $L_{\max}$ and the regrow cooldown), and our defaults were chosen on the same tasks that we report. The final grid size depends on where in the growth and pruning cycle training stops, so we report the average over the last growth interval. The grid schedule is the same for all tasks and seeds in our runs, which means SGQN does not currently adapt the resolution to the task. The ablations use 4 tasks with 5 seeds and the sensitivity study 2 tasks with 3 seeds, so small differences between variants are not resolved. The independence assumption of decoupled Q-learning (Eq.~\ref{eq:decqn}) can also be limiting for tasks with strongly coupled action dimensions. We report training returns over the final 100 episodes rather than separate evaluation episodes. The baseline comparison has further limits. CQN learns from pixels and the other methods from the state, so it is not a like-for-like baseline. \decqn\ and CQN were run with the settings stated in their publications and the defaults of our code where these are not stated, were not tuned, and reach much lower returns than the other methods, which we did not investigate. Some baseline runs produced no data (Table~\ref{tab:final_performance}), including all seven Bang-Bang PPO runs on Reacher Hard. We ran no significance tests. SGQN performs one gradient update per environment step and our \gqn\ and \decqn\ runs one per eight, so part of the advantage of SGQN may come from more training per sample. We did not run \gqn\ with the update frequency, exploration rule or network of SGQN, so the advantage over \gqn\ cannot be attributed to a specific design choice.

\paragraph{Future directions.} Several extensions are promising: (1) a growth rule that uses a measure of Q-value differences that is not dominated by their mean, for example the advantage of a bin over the dimension mean in units of its standard deviation, so that growth can actually discriminate between bins, (2) learning per-dimension growth schedules, (3) combining selective discretisation with model-based planning, and (4) applying SGQN to real robotic systems where sample efficiency and discrete actuation align naturally.

\section{Conclusion}
\label{sec:conclusion}

We introduced Selective Growing Q-Networks (SGQN), a value-based continuous control algorithm with an adaptive action grid and $\varepsilon$-greedy selection with a decaying exploration rate. SGQN reaches the highest mean final return on four of eight DeepMind Control Suite tasks, is within 1.1\% of the best method on three more, and is 3.4\% below Bang-Bang MPO on Walker Run, using on average 7.5 of 9 available bins per dimension. A direct analysis of the grid and a set of ablations show that this performance does not come from the adaptive grid: growth is effectively uniform, the Q-values of retained and removed bins differ by about one percent, and removing pruning changes the return by at most 1.1\%. What matters within SGQN in our experiments is starting from a coarse grid, which prevents occasional collapsed seeds, and $\varepsilon$-greedy selection, which an annealed softmax does not match. The growth and pruning hyperparameters can be varied widely without changing the return. We therefore read SGQN as a competitive decoupled Q-learner whose advantage over the baselines we cannot attribute to a single component, since it also differs from them in update frequency and other settings, and a growth rule using value differences between bins, which our rule does not exploit, remains an open design problem.

\appendix

\section{Full Hyperparameter Tables}
\label{app:hyperparams}

\begin{table}[H]
\centering
\caption{SGQN-specific hyperparameters. All values are passed to the training script as explicit command line options (initial bins, pruning interval, growth interval, minimum age, maximum lives, pruning threshold, growth threshold, regrow cooldown, selection strategy and $\varepsilon$ decay); the remaining options keep the defaults of our code.}
\label{tab:sgqn_hyperparams}
\begin{tabular}{ll}
\toprule
Hyperparameter & Value \\
\midrule
Full grid bins $N$ & 9 \\
Initial bins $n_0$ & 3 (boundary and centre bins, which are protected anchors) \\
Growth interval $\Delta_g$ & 300 episodes \\
Pruning interval $\Delta_p$ & 50 episodes \\
Growth threshold $\tau_Q$ & 0.85 (fraction of the best running mean Q) \\
Pruning threshold $\tau_p$ & 0.5 (quantile of the mature bins) \\
Minimum visits $v_{\min}$ & 2000 selections since activation \\
Maximum lives $L_{\max}$ & 1 pruning event \\
Minimum age $a_{\min}$ & 1 pruning event \\
Regrow cooldown $c$ & 3 pruning events \\
Minimum active bins $n_{\min}$ & 3 \\
Warm-up $T_\mathrm{warm}$ & 50 episodes \\
Exploration $\varepsilon$ & 0.1, multiplied by 0.995 per episode, minimum 0.01 \\
Selection (ablation only) & annealed softmax \\
\bottomrule
\end{tabular}
\end{table}

\begin{table}[H]
\centering
\caption{Network and optimisation hyperparameters of SGQN. The baselines use the settings stated in their original publications and the defaults of our code where a publication does not state a value (Table~\ref{tab:components}).}
\begin{tabular}{ll}
\toprule
Hyperparameter & Value \\
\midrule
Optimiser & Adam \\
Learning rate & $1 \times 10^{-4}$ \\
Discount $\gamma$ & 0.99 \\
Replay buffer size & $10^6$ \\
Batch size & 256 \\
$n$-step returns & 3 \\
Target update period & 100 gradient updates (one update per environment step) \\
Prioritised replay $\alpha$ & 0.6 \\
Prioritised replay $\beta$ & 0.4 (constant) \\
Gradient clip norm & 40.0 \\
Hidden layers & 2 $\times$ 512 \\
Activation & ELU \\
Normalisation & Layer norm \\
\bottomrule
\end{tabular}
\end{table}

\section{Environment Details}
\label{app:envs}

All methods except CQN use the standard DMControl state-based observation protocol with no frame stacking. CQN learns from 84$\times$84 images with 3 stacked frames and an action repeat of 2, as in its official configuration. Episodes run for 1000 environment steps. The random seeds are 0, 11, 21, 37, 42, 73 and 101, and they control both the environment initialisation and the neural network initialisation. No domain randomisation is applied.

\begin{table}[H]
\centering
\caption{DeepMind Control Suite task details.}
\begin{tabular}{lcc}
\toprule
Task & Action Dim. & State Dim. \\
\midrule
Reacher Easy  & 2  & 7  \\
Reacher Hard  & 2  & 7  \\
Hopper Hop    & 4  & 15 \\
Fish Swim     & 5  & 26 \\
Cheetah Run   & 6  & 17 \\
Walker Walk   & 6  & 24 \\
Walker Run    & 6  & 24 \\
Quadruped Walk & 12 & 78 \\
\bottomrule
\end{tabular}
\end{table}


\begin{thebibliography}{23}
\providecommand{\natexlab}[1]{#1}
\providecommand{\url}[1]{\texttt{#1}}
\expandafter\ifx\csname urlstyle\endcsname\relax
  \providecommand{\doi}[1]{doi: #1}\else
  \providecommand{\doi}{doi: \begingroup \urlstyle{rm}\Url}\fi

\bibitem[Abdolmaleki et~al.(2018)Abdolmaleki, Springenberg, Tassa, Munos,
  Heess, and Riedmiller]{abdolmaleki2018maximum}
Abbas Abdolmaleki, Jost~Tobias Springenberg, Yuval Tassa, R{\'e}mi Munos,
  Nicolas Heess, and Martin Riedmiller.
\newblock Maximum a posteriori policy optimisation.
\newblock \emph{arXiv preprint arXiv:1806.06920}, 2018.

\bibitem[Ba et~al.(2016)Ba, Kiros, and Hinton]{ba2016layer}
Jimmy~Lei Ba, Jamie~Ryan Kiros, and Geoffrey~E Hinton.
\newblock Layer normalization.
\newblock \emph{arXiv preprint arXiv:1607.06450}, 2016.

\bibitem[Coulom(2007)]{coulom2007computing}
R{\'e}mi Coulom.
\newblock Computing elo ratings of move patterns in the game of go.
\newblock In \emph{Computer Games Workshop}, 2007.

\bibitem[Czarnecki et~al.(2018)Czarnecki, Jayakumar, Jaderberg, Hasenclever,
  Teh, Heess, Osindero, and Pascanu]{czarnecki2018mix}
Wojciech~M Czarnecki, Siddhant~M Jayakumar, Max Jaderberg, Leonard Hasenclever,
  Yee~Whye Teh, Nicolas Heess, Simon Osindero, and Razvan Pascanu.
\newblock Mix and match -- agent curricula for reinforcement learning.
\newblock In \emph{International Conference on Machine Learning}, 2018.

\bibitem[Fujimoto et~al.(2018)Fujimoto, van Hoof, and
  Meger]{fujimoto2018addressing}
Scott Fujimoto, Herke van Hoof, and David Meger.
\newblock Addressing function approximation error in actor-critic methods.
\newblock In \emph{International Conference on Machine Learning}, 2018.

\bibitem[Gu et~al.(2016)Gu, Lillicrap, Sutskever, and Levine]{gu2016continuous}
Shixiang Gu, Timothy Lillicrap, Ilya Sutskever, and Sergey Levine.
\newblock Continuous deep q-learning with model-based acceleration.
\newblock In \emph{International Conference on Machine Learning}, 2016.

\bibitem[Haarnoja et~al.(2018)Haarnoja, Zhou, Abbeel, and
  Levine]{haarnoja2018soft}
Tuomas Haarnoja, Aurick Zhou, Pieter Abbeel, and Sergey Levine.
\newblock Soft actor-critic: Off-policy maximum entropy deep reinforcement
  learning with a stochastic actor.
\newblock In \emph{International Conference on Machine Learning}, 2018.

\bibitem[Lillicrap et~al.(2016)Lillicrap, Hunt, Pritzel, Heess, Erez, Tassa,
  Silver, and Wierstra]{lillicrap2016continuous}
Timothy~P Lillicrap, Jonathan~J Hunt, Alexander Pritzel, Nicolas Heess, Tom
  Erez, Yuval Tassa, David Silver, and Daan Wierstra.
\newblock Continuous control with deep reinforcement learning.
\newblock In \emph{International Conference on Learning Representations}, 2016.

\bibitem[Mnih et~al.(2015)Mnih, Kavukcuoglu, Silver, Rusu, Veness, Bellemare,
  Graves, Riedmiller, Fidjeland, Ostrovski, et~al.]{mnih2015human}
Volodymyr Mnih, Koray Kavukcuoglu, David Silver, Andrei~A Rusu, Joel Veness,
  Marc~G Bellemare, Alex Graves, Martin Riedmiller, Andreas~K Fidjeland, Georg
  Ostrovski, et~al.
\newblock Human-level control through deep reinforcement learning.
\newblock \emph{Nature}, 518\penalty0 (7540):\penalty0 529--533, 2015.

\bibitem[Rashid et~al.(2018)Rashid, Samvelyan, Schroeder, Farquhar, Foerster,
  and Whiteson]{rashid2018qmix}
Tabish Rashid, Mikayel Samvelyan, Christian Schroeder, Gregory Farquhar, Jakob
  Foerster, and Shimon Whiteson.
\newblock {QMIX}: Monotonic value function factorisation for decentralised
  multi-agent reinforcement learning.
\newblock In \emph{International Conference on Machine Learning}, 2018.

\bibitem[Schaul et~al.(2016)Schaul, Quan, Antonoglou, and
  Silver]{schaul2016prioritized}
Tom Schaul, John Quan, Ioannis Antonoglou, and David Silver.
\newblock Prioritized experience replay.
\newblock In \emph{International Conference on Learning Representations}, 2016.

\bibitem[Schulman et~al.(2017)Schulman, Wolski, Dhariwal, Radford, and
  Klimov]{schulman2017proximal}
John Schulman, Filip Wolski, Prafulla Dhariwal, Alec Radford, and Oleg Klimov.
\newblock Proximal policy optimization algorithms.
\newblock \emph{arXiv preprint arXiv:1707.06347}, 2017.

\bibitem[Seo et~al.(2024)Seo, Uru{\c{c}}, and James]{seo2024continuous}
Younggyo Seo, Jafar Uru{\c{c}}, and Stephen James.
\newblock Continuous control with coarse-to-fine reinforcement learning.
\newblock In \emph{Advances in Neural Information Processing Systems}, 2024.

\bibitem[Seyde et~al.(2021)Seyde, Gilitschenski, Schwarting, Stellato,
  Riedmiller, Wulfmeier, and Rus]{seyde2021bangbang}
Tim Seyde, Igor Gilitschenski, Wilko Schwarting, Bartolomeo Stellato, Martin
  Riedmiller, Markus Wulfmeier, and Daniela Rus.
\newblock Is bang-bang control all you need? solving continuous control with
  {B}ernoulli policies.
\newblock In \emph{Advances in Neural Information Processing Systems}, 2021.

\bibitem[Seyde et~al.(2022)Seyde, Werner, Schwarting, Gilitschenski,
  Riedmiller, Rus, and Wulfmeier]{seyde2022solving}
Tim Seyde, Peter Werner, Wilko Schwarting, Igor Gilitschenski, Martin
  Riedmiller, Daniela Rus, and Markus Wulfmeier.
\newblock Solving continuous control via q-learning.
\newblock \emph{arXiv preprint arXiv:2210.12566}, 2022.

\bibitem[Seyde et~al.(2024)Seyde, Werner, Schwarting, Wulfmeier, and
  Rus]{seyde2024growing}
Tim Seyde, Peter Werner, Wilko Schwarting, Markus Wulfmeier, and Daniela Rus.
\newblock Growing {Q}-networks: Solving continuous control tasks with adaptive
  control resolution.
\newblock In \emph{International Conference on Learning Representations}, 2024.

\bibitem[Sunehag et~al.(2018)Sunehag, Lever, Gruslys, Czarnecki, Zambaldi,
  Jaderberg, Lanctot, Sonnerat, Leibo, Tuyls, et~al.]{sunehag2018value}
Peter Sunehag, Guy Lever, Audrunas Gruslys, Wojciech~Marian Czarnecki, Vinicius
  Zambaldi, Max Jaderberg, Marc Lanctot, Nicolas Sonnerat, Joel~Z Leibo, Karl
  Tuyls, et~al.
\newblock Value-decomposition networks for cooperative multi-agent learning.
\newblock In \emph{International Conference on Autonomous Agents and
  Multi-Agent Systems}, 2018.

\bibitem[Sutton \& Barto(2018)Sutton and Barto]{sutton2018reinforcement}
Richard~S Sutton and Andrew~G Barto.
\newblock \emph{Reinforcement Learning: An Introduction}.
\newblock MIT Press, 2nd edition, 2018.

\bibitem[Tang \& Agrawal(2020)Tang and Agrawal]{tang2020discretising}
Yunhao Tang and Shipra Agrawal.
\newblock Discretizing continuous action space for on-policy optimization.
\newblock In \emph{AAAI Conference on Artificial Intelligence}, 2020.

\bibitem[Tassa et~al.(2018)Tassa, Tunyasuvunakool, Muldal, Doron, Liu, Bohez,
  Merel, Erez, Lillicrap, and Heess]{tassa2018deepmind}
Yuval Tassa, Saran Tunyasuvunakool, Alistair Muldal, Yotam Doron, Siqi Liu,
  Steven Bohez, Josh Merel, Tom Erez, Timothy Lillicrap, and Nicolas Heess.
\newblock {DeepMind} control suite.
\newblock \emph{arXiv preprint arXiv:1801.00690}, 2018.

\bibitem[Tavakoli et~al.(2018)Tavakoli, Pardo, and
  Kormushev]{tavakoli2018action}
Arash Tavakoli, Fabio Pardo, and Petar Kormushev.
\newblock Action branching architectures for deep reinforcement learning.
\newblock In \emph{AAAI Conference on Artificial Intelligence}, 2018.

\bibitem[van Hasselt et~al.(2016)van Hasselt, Guez, and
  Silver]{vanHasselt2016deep}
Hado van Hasselt, Arthur Guez, and David Silver.
\newblock Deep reinforcement learning with double q-learning.
\newblock In \emph{AAAI Conference on Artificial Intelligence}, 2016.

\bibitem[Watkins \& Dayan(1992)Watkins and Dayan]{watkins1992q}
Christopher~JCH Watkins and Peter Dayan.
\newblock Q-learning.
\newblock \emph{Machine Learning}, 8\penalty0 (3-4):\penalty0 279--292, 1992.

\end{thebibliography}
\end{document}